\documentclass[conference]{IEEEtran}
\IEEEoverridecommandlockouts
\usepackage{cite}
\usepackage{amsmath,amssymb,amsfonts}
\usepackage{algorithmic}
\usepackage{graphicx}
\usepackage{textcomp}
\usepackage{xcolor}
\usepackage{url}
\usepackage{booktabs} 
\usepackage{float}    
\usepackage{subfig}   
\usepackage{placeins} 

\def\BibTeX{{\rm B\kern-.05em{\sc i\kern-.025em b}\kern-.08em
    T\kern-.1667em\lower.7ex\hbox{E}\kern-.125emX}}
\begin{document}

\title{Accelerating Local AI on Consumer GPUs: A Hardware-Aware Dynamic Strategy for YOLOv10s}

\author{%
\IEEEauthorblockN{Mahmudul Islam Masum}
\IEEEauthorblockA{\textit{School of Computing and}\\ \textit{Information Sciences}\\
\textit{Florida International University}\\
Miami, FL, USA\\
mmasu004@fiu.edu}
\and
\IEEEauthorblockN{Miad Islam}
\IEEEauthorblockA{\textit{Department of Computer and}\\ \textit{Information Systems}\\
\textit{Saint Leo University}\\
Tampa, FL, USA\\
miad.islam@saintleo.edu}
}

\maketitle

\begin{abstract}
As local AI grows in popularity, there is a critical gap between the benchmark performance of object detectors and their practical viability on consumer-grade hardware. While models like YOLOv10s promise real-time speeds, these metrics are typically achieved on high-power, desktop-class GPUs. This paper reveals that on resource-constrained systems, such as laptops with RTX 4060 GPUs, performance is not compute-bound but is instead dominated by system-level bottlenecks, as illustrated by a simple bottleneck test. To overcome this hardware-level constraint, we introduce a Two-Pass Adaptive Inference algorithm, a model-independent approach that requires no architectural changes. This study mainly focuses on ‘adaptive’ inference strategies and undertakes a comparative analysis of architectural early-exit and resolution-adaptive routing, highlighting their respective trade-offs within a unified evaluation framework. The system uses a fast, low-resolution pass and only escalates to a high-resolution model pass when detection confidence is low. On a 5000-image COCO dataset, our method achieves a 1.85x speedup over a PyTorch Early-Exit baseline, with a modest mAP loss of 5.51\%. This work provides a practical and reproducible blueprint for deploying high-performance, real-time AI on consumer-grade devices by shifting the focus from pure model optimization to hardware-aware inference strategies that maximize throughput.
\end{abstract}


\section{\textbf{Introduction}}
Object detection has become one of the foundations of computer vision. It's behind everything from industrial automation and robotics to augmented reality apps. Deep learning has pushed the field forward dramatically, and among the many models out there, the YOLO family (You Only Look Once) has stood out for balancing speed with surprisingly strong accuracy. One of the latest versions, YOLOv10s, continues in that direction and is increasingly seen as a go-to choice for real-time tasks.

However, there’s a gap between how these models are typically evaluated and where they’re deployed. Most of the benchmarks are done on powerful, server-grade GPUs like the NVIDIA A100, which has plenty of cooling and massive power-draw. But in the real-world scenario, object detection is often executed on laptops, mobile devices, and embedded systems, where thermal limits and power limits can significantly slow things down.

This mismatch affects developers trying to build responsive apps for everyday devices. Inference times can vary wildly, and user experience often takes a hit. And in scenarios where privacy matters, for example, in home security systems or autonomous navigation, we can’t always count on the cloud. Local inference becomes essential. Most optimization strategies still tend to prioritize reducing FLOPs or shrinking model size. Those strategies work, but they don’t necessarily fix the broader system-level bottlenecks in constrained environments.

That’s the gap we’re trying to address here. Instead of focusing only on algorithmic tweaks, our research takes a step back to asking a different kind of question: How well do these modern detection models hold up on the kind of hardware people use every day? We targeted the RTX 4060 Laptop GPU which is not top of the line, but not underpowered either. It’s a good example of what a mid-range user might have access to. It also supports acceleration features like half-precision (FP16) inference, while still dealing with strict power and thermal constraints, which makes it an ideal real-world test case.

To work within those limitations, we introduce a lightweight, adaptive approach that adjusts computing effort depending on how difficult a frame appears to be. This is a two-pass adaptive inference algorithm that uses the standard YOLOv10s model at different resolutions. The system first runs a fast, low-resolution pass and only escalates to a high-resolution pass if the initial confidence is low. This approach is contrasted with an architectural early-exit (EE) baseline, which is a common strategy for dynamic inference. Our work presents a hardware-aware strategy that not only enhances object detection performance but also provides a conceptual framework for managing computational loads in any local AI task, ensuring a more responsive and practical user experience. This study can also be seen as a comparative study between two adaptive inference strategies, highlighting the trade-offs of different approaches. 

\section{\textbf{Literature Review}}
Real-time object detection has come a long way in the past decade, mostly led by the YOLO (You Only Look Once) models. The YOLOv10 end-to-end (E2E) variant removes the need for non-maximum suppression (NMS) and improves latency by adopting an end-to-end architecture \cite{b1}. This makes it well-suited for real-time applications, especially on devices with limited computational power. However, most of the testing and validation still happens on high-end GPUs, which don’t really reflect the kinds of mobile or embedded environments these models are increasingly expected to be used in.

Meanwhile, dynamic neural networks have been gaining ground, particularly those that feature early-exit mechanisms \cite{b2}. These models adapt inference effort based on the complexity of the input, allowing for quicker exits on easier examples. It's a strategy that shares some similarity with our two-pass approach, which also aims to preserve computing power where and when it's not needed. Another closely related idea comes from Yang et al., whose Resolution Adaptive Network (RANet) routes simpler images through a lightweight, low-res subnetwork, only escalating to more expensive computation when necessary \cite{b3}. That concept directly supports our confidence-triggered fallback design.

While YOLO has been shown to be effective for detection under challenging microscopic conditions \cite{b4}, in more specific domains, other techniques have also emerged. AEA-YOLO \cite{b5} and GOIS \cite{b6}, for example, have shown strong results in tougher conditions including foggy scenes, and tiny object detection. But these systems often depend on additional modules or require architectural redesigns and specialized retraining, which can make them harder to integrate. However, our approach adds a lightweight control layer on top of a standard architecture, making it simpler to deploy.

Broader surveys in this field have confirmed a recurring challenge, even as detection models improve, inference speed often runs up against system-level bottlenecks, things like thermal throttling or limited memory I/O bandwidth, especially on edge hardware \cite{b7}. In this context, strategies like Adaptive Feeding \cite{b8}, which choose between fast and accurate models at runtime, have shown that inference doesn’t have to be one-size-fits-all. Our work is very much in the same spirit, but we take it a step further: we remove the need for multiple models or retraining altogether, simplifying the deployment process.

There’s also a growing awareness about reliability, privacy, and accessibility. In many real-world cases, like smart home or healthcare, relying on cloud connectivity isn’t always feasible, and sending visual data off-device raises valid privacy concerns. Local inference on consumer-grade GPUs addresses both. Nothing leaves the device, which offers a layer of privacy that’s becoming more of a baseline expectation than an extra feature.

Finally, it’s worth noting that early-exit architectures \cite{b2} and dynamic inference strategies \cite{b9} indirectly support this current trend of smarter, adaptive edge-computing. These methods don’t just make models faster, they make them more context-aware and better suited for real-world use.

\section{\textbf{Methodology}}
\begin{figure}[h]
\centering
\includegraphics[width=\linewidth]{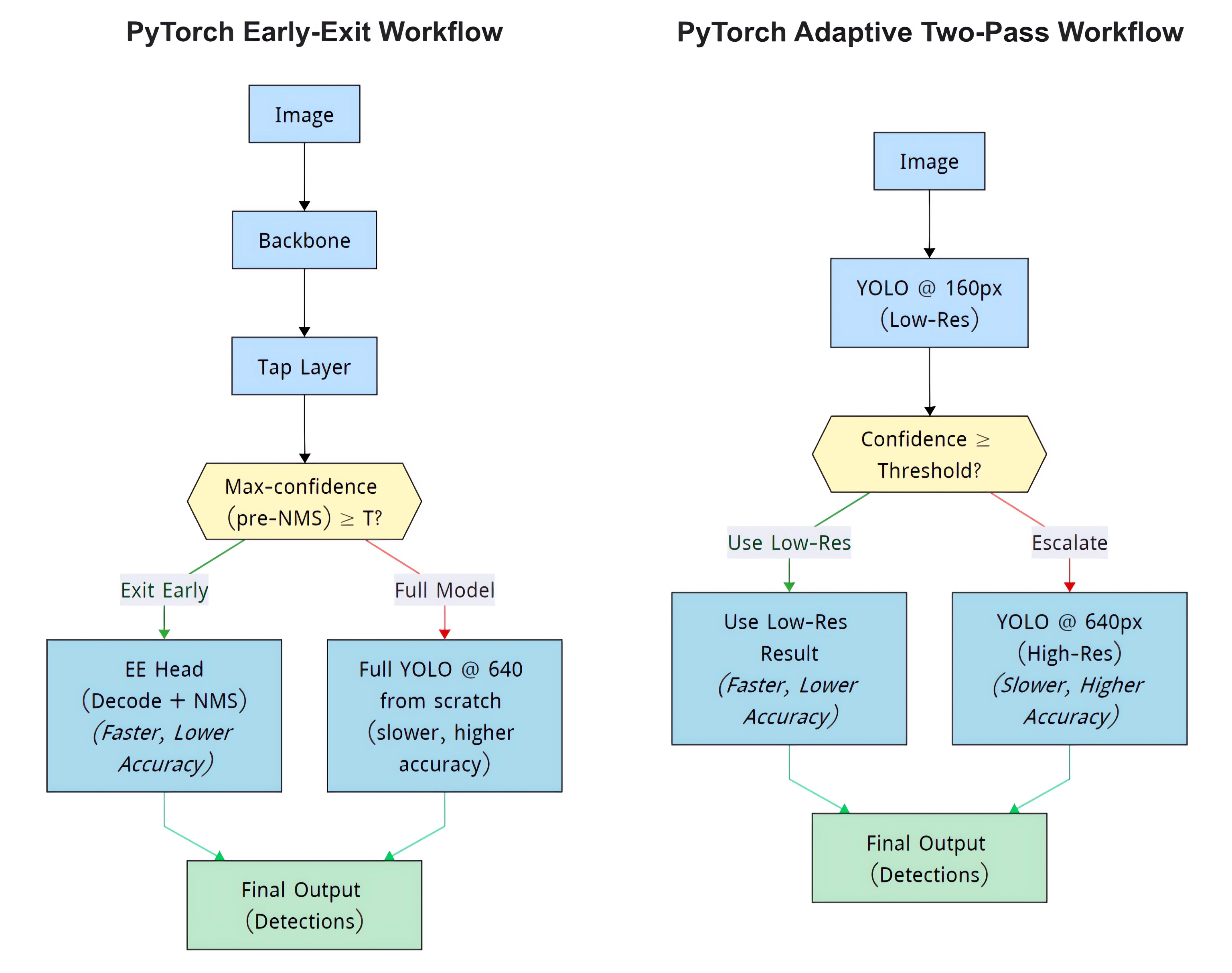}
\caption{Conceptual workflows of the PyTorch Early-Exit and PyTorch Adaptive Two-Pass strategies.}
\label{fig:workflow}
\end{figure}

In this study, we evaluate real-time object detection performance using YOLOv10s on consumer-grade hardware. The NVIDIA RTX 4060 Laptop GPU, with 3072 CUDA cores and support for FP16 precision, strikes a balance between capability and constraint. It's not a flagship chip, but it’s a good example of a power- and a thermally-constrained environment. The test system used an Intel Core i7-13700HX CPU, paired with 8GB of VRAM and 48GB of DDR5 RAM. On the software side, we used Python 3.11 and PyTorch 2.1.2 (compiled with CUDA 12.1). The YOLOv10s baseline model was loaded directly from its official PyTorch checkpoint. Although YOLOv10 includes an (E2E) variant without NMS, in this work we used the standard Ultralytics YOLOv10s implementation with NMS enabled, since it is the default and compatible with our Early-Exit head. All experiments leverage PyTorch's native automatic mixed-precision capabilities for FP16 acceleration on the GPU. For reproducibility, all experiments used Seed=123, the official Ultralytics YOLOv10s checkpoint (May 2024), and the full COCO 2017 validation set (5,000 images). Fig. \ref{fig:workflow} provides a high-level overview of the setup.

\subsection{Bottleneck Test}
To understand what was limiting inference speed, model size or hardware, we performed a simple bottleneck test. The idea was straightforward; we compared performance using two drastically different input resolutions. In Table \ref{tab:bottleneck}, we can see, even when we reduced the input size by a factor of 100, the gain in FPS was surprisingly low. That result suggests the bottleneck isn’t just the model’s computational demand, but rather system-level constraints like memory bandwidth and power delivery.
\begin{table}[H]
\centering
\caption{Bottleneck Test Results.}
\label{tab:bottleneck}
\begin{tabular}{@{}lc@{}}
\toprule
\textbf{Input Size} & \textbf{FPS} \\
\midrule
640$\times$640 & 21.43 \\
64$\times$64 & 38.63 \\
\bottomrule
\end{tabular}
\end{table}
This aligns with previous findings in bottleneck-aware inference \cite{b7}, \cite{b10}. While a conclusive diagnosis would require more extensive monitoring, this result suggests that performance on this consumer-grade hardware is not purely compute-bound. A more detailed visual analysis of the hardware’s performance is available in the Results section. 

\subsection{Dynamic Inference Strategies}
We implemented and compared two distinct dynamic inference methods using a standard PyTorch-based YOLOv10s model. The first was an architectural Early-Exit model to serve as a strong baseline, and the second was our proposed Two-Pass Adaptive Inference algorithm.

\subsubsection{Early-Exit (EE) Baseline}
To create a competitive baseline for comparison, we implemented an architectural early-exit mechanism in PyTorch. A lightweight detection head (\texttt{EarlyExitHead}) was attached to an intermediate feature map of the YOLOv10s backbone (specifically, tap layer 16). For a given input, if the confidence score produced by this early head exceeded a tuned threshold, inference would terminate. Otherwise, computation would continue through the full, high-resolution model. To maximize the accuracy of this baseline, Test-Time Augmentation (TTA) via a horizontal flip was also applied to all images processed through the early-exit path.

\subsubsection{Two-Pass Adaptive Inference}
Our proposed method is a model-agnostic runtime algorithm that requires no architectural modifications. Its design was inspired by established concepts in dynamic inference, including early-exit architectures, resolution-adaptive networks, and adaptive feeding strategies \cite{b2, b3, b6, b8}. The model's operation can be described in three stages. It begins with a first pass, where an input image is processed by the model at a fast, low-resolution (160x160). Following this, a gating step extracts the maximum confidence score from all resulting detections. Finally, this score is used for escalation; if it is below a tuned confidence threshold, the system processes the original image again using the same model at its more accurate, high-resolution (640x640). If the confidence is sufficient, the low-resolution results are accepted.
This approach dynamically allocates computational resources based on inference confidence, saving significant time on inputs where the model is already certain after a cheap initial pass. Unless noted, we used score threshold 0.001 and NMS IoU 0.70 for post-processing; the Early-Exit path used a single-flip TTA; all inference ran in FP16 on CUDA.

\subsection{Threshold Calibration and Tuning}
A critical component of both dynamic strategies is the selection of the confidence threshold. As implemented in the benchmark script, a naive or fixed threshold is suboptimal, so we used a rigorous two-stage tuning process. The process began with an initial calibration on a 400-image subset to find a "seed" threshold based on a predefined target rate, such as a 50\% exit rate for the Early-Exit baseline. This seed value was then refined through a performance-oriented grid search on a separate 200-image tuning subset. The optimization goal was tailored to each model for a fair comparison. For the Early-Exit baseline, the final threshold was selected to maximize the mean average precision (mAP) while maintaining at least a 10\% early exit rate. In contrast, the threshold for the Two-Pass Adaptive model was chosen to find an optimal balance between inference speed and accuracy relative to the EE baseline's performance. This comprehensive tuning ensures that both models were configured to their optimal operating points before the final benchmark.

\subsection{Dataset and Categorization}
\begin{figure}[h]
\centering
\subfloat[Examples of simple images.\label{fig:simple_examples}]{
\includegraphics[width=0.48\linewidth]{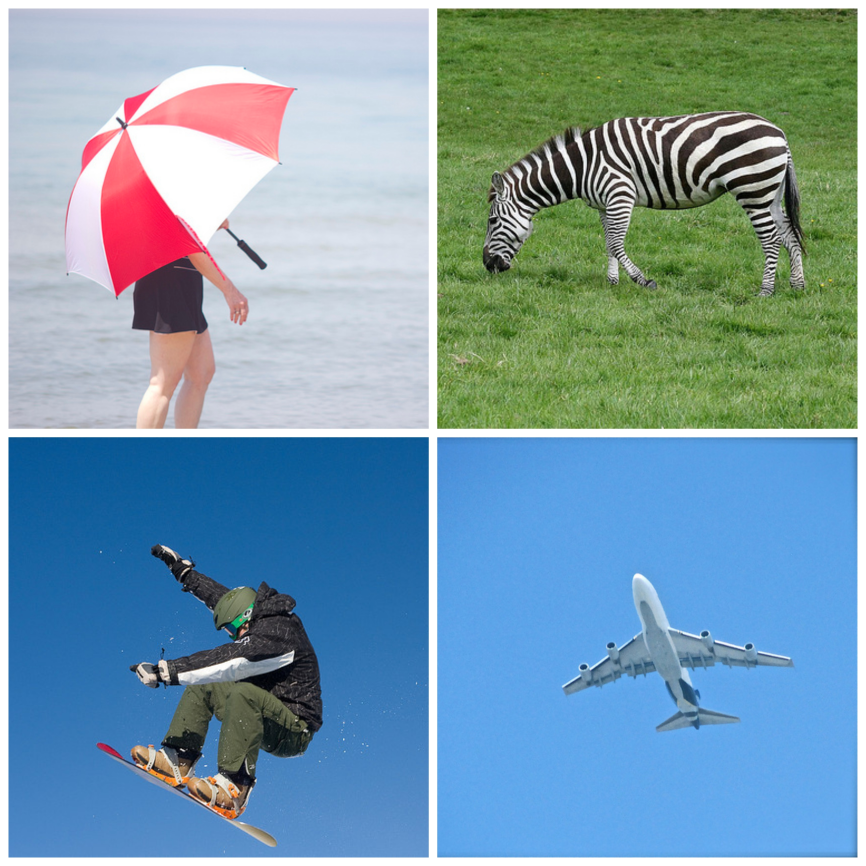}
}
\hfill
\subfloat[Examples of complex images.\label{fig:complex_examples}]{
\includegraphics[width=0.48\linewidth]{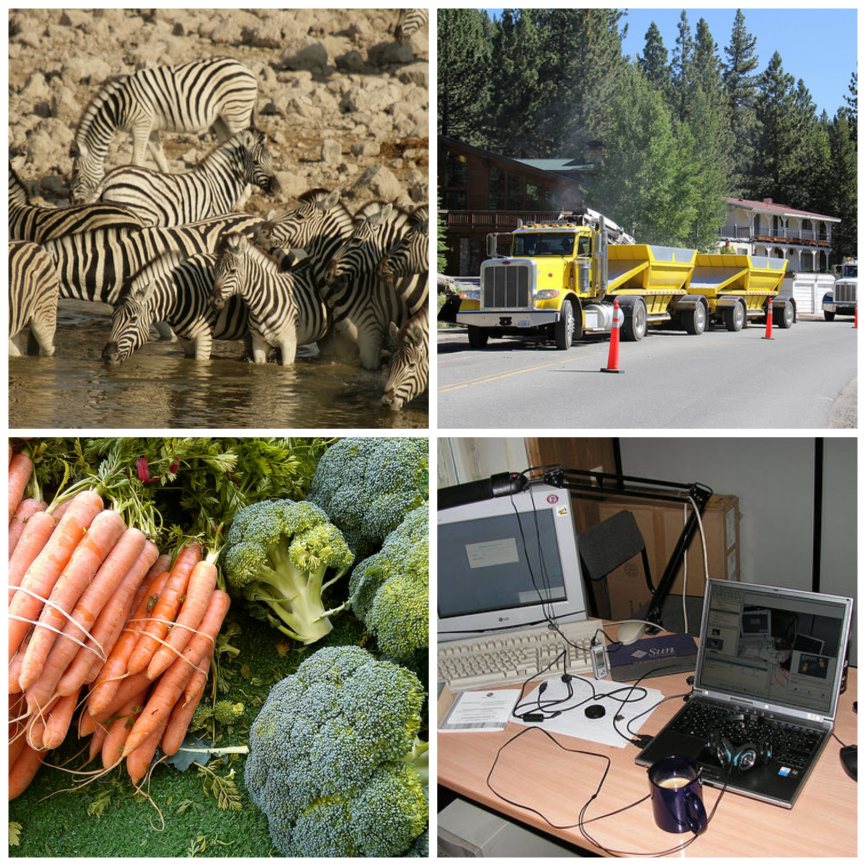}
}
\caption{Image categorization based on object count and visual clutter.}
\label{fig:image_examples}
\end{figure}

We additionally created a balanced subset of 2,500 COCO images (1,250 ‘simple’ and 1,250 ‘complex’), chosen to approximate real-world variation. This was used as a demonstration to show how the two-pass system behaves under controlled conditions. The trends matched the main benchmark results, confirming the practical value of the approach. A more detailed illustration of the adaptive system’s behavior can be found in the Results and Analysis section (see Fig. 7). This qualitative subset helps to illustrate how adaptive inference decisions unfold at both low and high confidence levels. It also provides an intuitive perspective that complements the quantitative benchmarks reported on the full validation set. 

The final benchmark was conducted on the entire 5000-image COCO 2017 validation set to provide a more standardized and reproducible result. We report official metrics only on the full 5,000-image COCO-val set; the balanced subset is purely demonstrative.

\section{\textbf{Results and Analysis}}
\begin{figure}[!h]
\centering
\includegraphics[width=\linewidth]{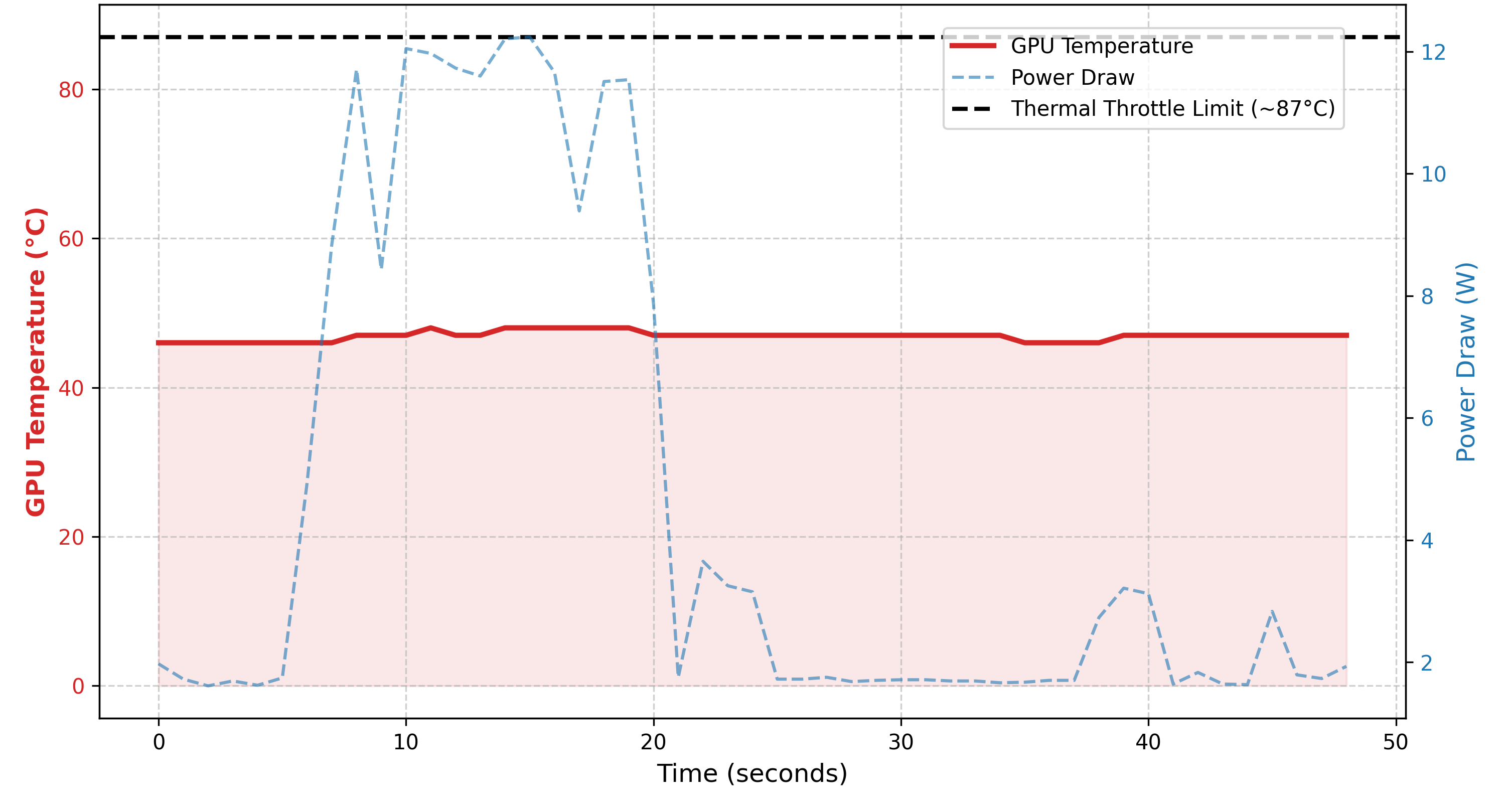}
\caption{GPU power consumption patterns.}
\label{fig:bottleneck_graph}
\end{figure}

To understand what was really going on, we looked more closely at the system-level behavior through our Bottleneck Test (see Fig. \ref{fig:bottleneck_graph}). While the GPU temperature stayed well below its thermal throttle point (a common temperature of 87°C on many laptops), the power consumption told a different story. As seen in Fig. 3, it fluctuated heavily and rarely hit the GPU’s maximum power ceiling. This behavior suggests that the bottleneck is not a simple power or thermal limit, but a more complex system-level constraint. Even with a 100$\times$ reduction in input pixels, the system could not fully capitalize on the reduced computational load, indicating that factors such as memory I/O, driver overhead, or other platform limits were preventing further performance gains.

\begin{table}[!h]
\centering
\caption{Final Benchmark on COCO 2017 Validation Set}
\label{tab:final_benchmark}
\begin{tabular}{@{}lcccc@{}}
\toprule
\textbf{Model} & \textbf{it/s} & \textbf{Speedup} & \textbf{mAP}& \textbf{mAP Loss (\%)} \\
\midrule
PyTorch Early-Exit & 27.49 & 1.00x & 0.399& 0.00\% \\
Adaptive Two-Pass & 50.99 & 1.85x & 0.377& -5.51\%\\
\bottomrule
\end{tabular}
\end{table}
\begin{table}[!h]
\centering
\caption{Routing Information.}
\label{tab:routing_info}
\begin{tabular}{@{}lc@{}}
\toprule
\textbf{Model} & \textbf{Routing Info} \\
\midrule
PyTorch Early-Exit & 14.5\% Exited Early (Gate T = 0.491)\\
Adaptive Two-Pass & 28.1\% First Pass, 71.9\% Escalated (T = 0.859)\\
\bottomrule
\end{tabular}
\end{table}

Our final evaluation was conducted on the full 5000-image COCO 2017 validation set to provide a robust comparison against a PyTorch architectural early-exit (EE) baseline. The results, detailed in Table \ref{tab:final_benchmark}, demonstrate a significant performance improvement. Our adaptive pipeline achieved a processing speed of 50.99 it/s (iterations per second), resulting in a 1.85x speedup over the EE baseline's 27.49 it/s. This speed increase comes at the cost of a 5.51\% drop in mAP, with the adaptive system's mAP@[.5:.95] at 0.377 compared to the early-exit baseline's 0.399. This establishes a compelling case for our method's efficiency, where an 85\% speed gain is achieved for a modest reduction in accuracy.

\begin{figure}
\centering
\includegraphics[width=0.95\linewidth]{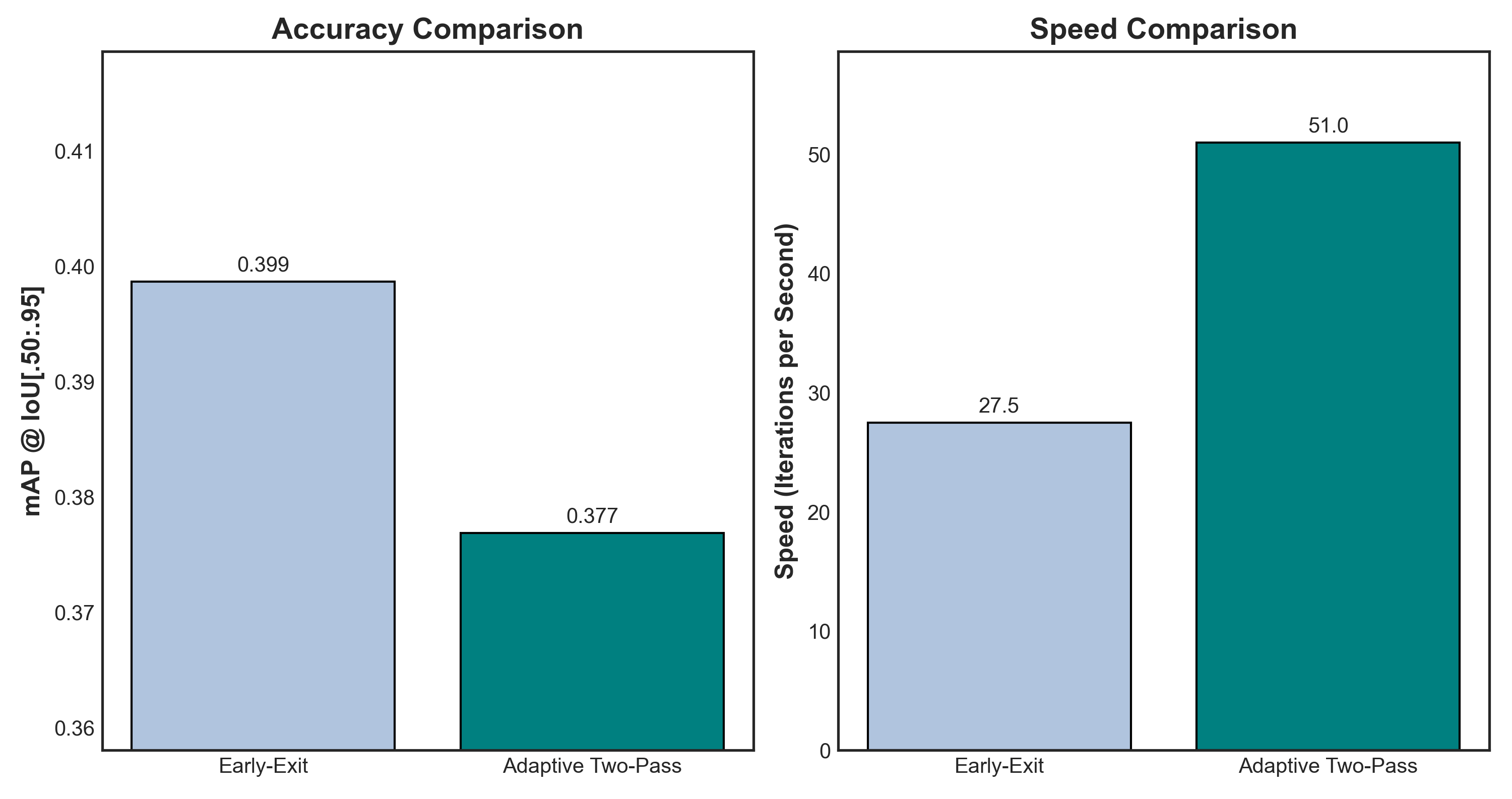}
\caption{Accuracy and speed comparison between the two models.}
\label{fig:4}
\end{figure}

Fig.~\ref{fig:4} provides a direct comparison between the Adaptive Two-Pass system and the Early-Exit baseline on the COCO 2017 validation set. The chart highlights that the Adaptive pipeline achieves 51.0 it/s, representing a 1.85$\times$ speedup over the Early-Exit model’s 27.5 it/s. In terms of accuracy, the Adaptive approach has an mAP of 0.377, compared to 0.399 for the more accurate but slower baseline. These results confirm the improvements reported in Table~\ref{tab:final_benchmark}.

\begin{figure}[!h]
\centering
\includegraphics[width=0.95\linewidth]{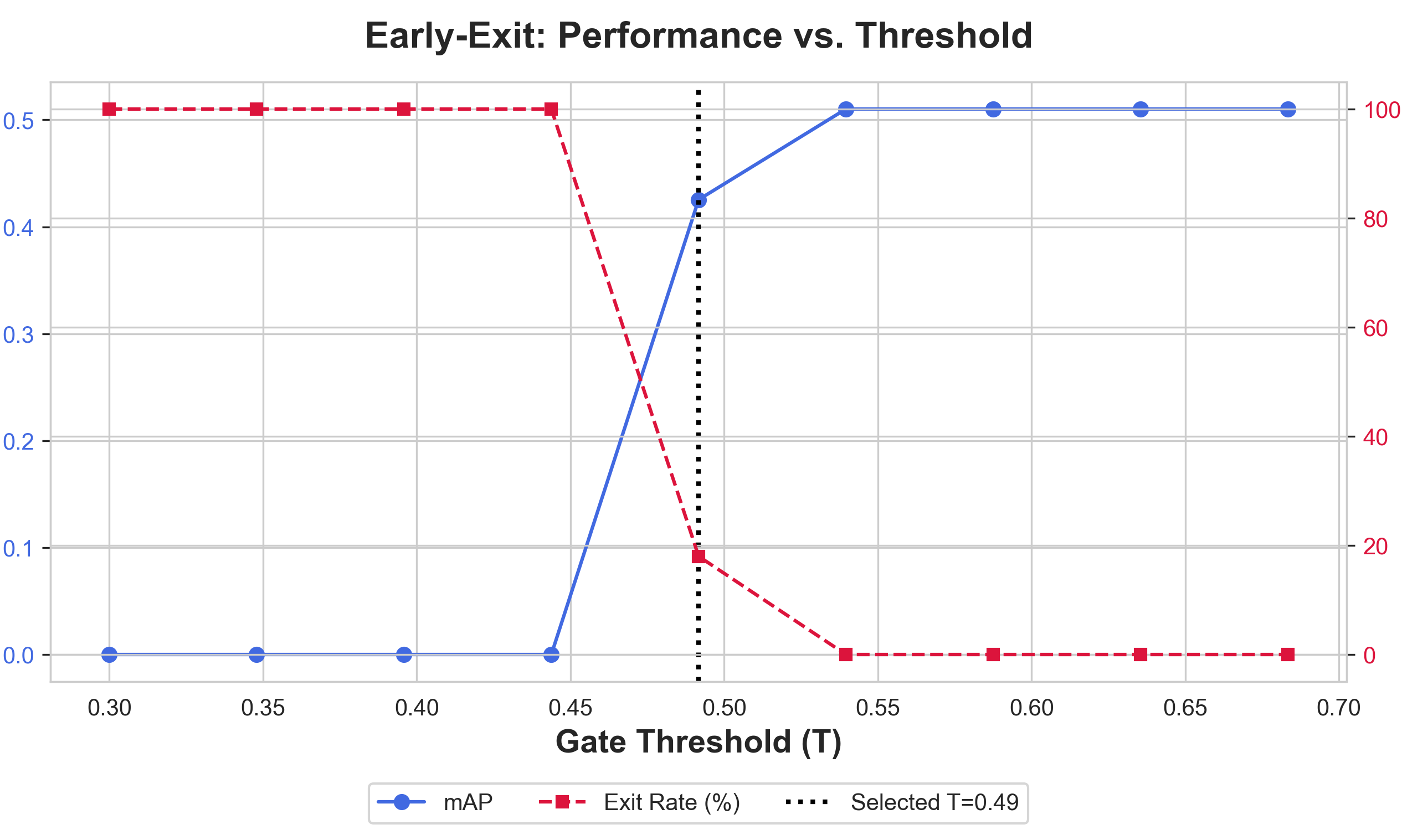}
\caption{Early-Exit trade-off between mAP and exit rate across different gate thresholds.}
\label{fig:5}
\end{figure}

Fig.~\ref{fig:5} shows the trade-off behavior of the Early-Exit model as the gate threshold varies. At low thresholds, nearly all inputs exit early, but the resulting mAP collapses. As the threshold increases, fewer images exit early and accuracy improves. The chosen operating point at $T \approx 0.49$ balances this trade-off, yielding an mAP of 0.399 while allowing 14.5\% of images to exit early. This setting was used in the final evaluation to ensure a fair comparison.

\begin{figure}[h]
\centering
\includegraphics[width=0.95\linewidth]{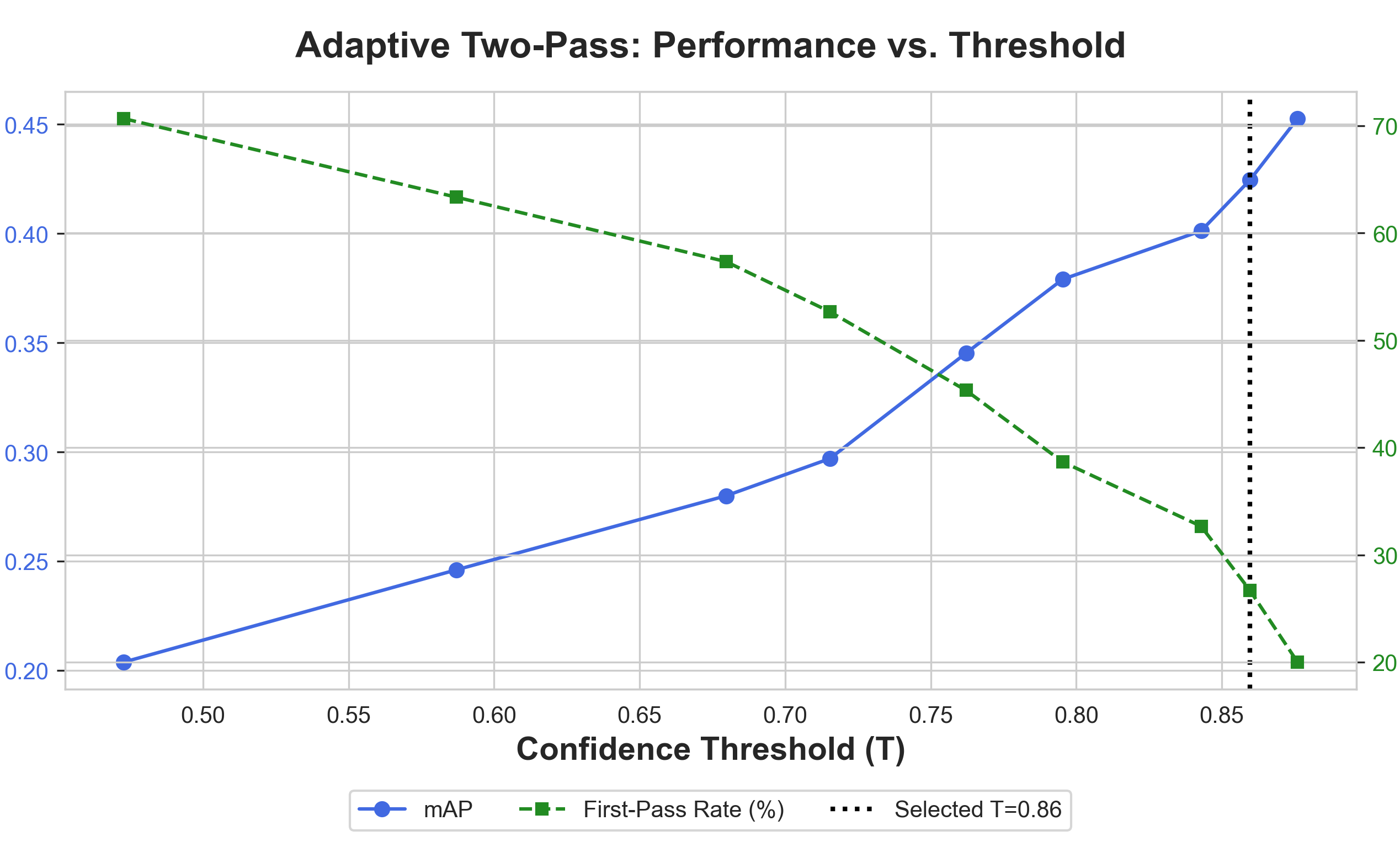}
\caption{Adaptive Two-Pass trade-off between mAP and first-pass rate across confidence thresholds.}
\label{fig:6}
\end{figure}

Fig.~\ref{fig:6} illustrates the trade-off for the Adaptive Two-Pass pipeline across different confidence thresholds. Lower thresholds increase the first-pass rate but reduce overall accuracy, while higher thresholds escalate more images to the second pass, improving mAP. At the selected threshold of $T \approx 0.86$, the Adaptive system achieves an mAP of 0.377, with 28.1\% of images resolved in the first pass and 71.9\% escalated. This operating point reflects the routing statistics reported in Table~\ref{tab:routing_info}.

\begin{figure}[H]
\centering
\includegraphics[width=0.95\linewidth]{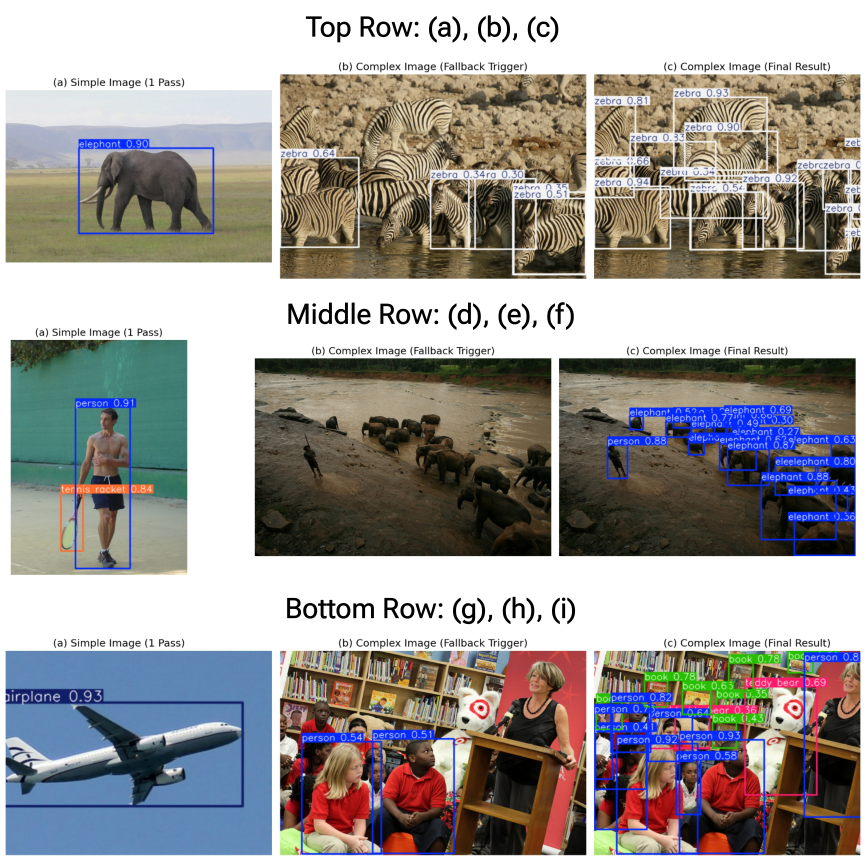}
\caption{Qualitative examples showing how the Adaptive Two-Pass system handles simple vs complex scenes.}
\label{fig:7}
\end{figure}

The qualitative results in Fig.~\ref{fig:7} give an intuitive sense of how the Adaptive Two-Pass mechanism operates in practice. In straightforward cases, such as the examples shown in (a, d, g), the system confidently detects the main object in a single low-resolution pass. For instance, the elephant is recognized immediately without requiring escalation. However, in more cluttered and visually complex scenes, such as (b, c, e, f, h, i), the first pass produces low-confidence outputs. These cases trigger the high-resolution second pass, which significantly improves detection accuracy. This behavior demonstrates that the system effectively allocates computational effort where it is needed most, saving resources on simpler inputs while applying additional computation to harder cases. This dynamic allocation of computational effort is a hallmark of adaptive inference strategies, a tactic similar to models designed for other complex scenarios like nighttime traffic detection \cite{b11}.

\section{\textbf{Discussion}}
Our findings confirm a core insight: for consumer GPUs like the RTX 4060, inference speed is often limited less by the model's computational complexity and more by system-level bottlenecks, potentially including memory bandwidth, host-to-device data transfer, or platform-specific power management policies \cite{b7}, \cite{b12}. Furthermore, our results highlight the practical trade-off between theoretical precision and deployable performance. Using native FP16 in PyTorch, while necessary for speed, incurs a notable drop in absolute accuracy, although the relative trends between Early-Exit and Adaptive remain consistent. This underscores the reality that deployment on resource-constrained hardware often involves navigating a complex balance of speed, precision, and power consumption. This reality is often overlooked in academic benchmarks performed on server-grade GPUs like the NVIDIA A100 \cite{b10}. Those machines are typically compute-bound, and they benefit more directly from reductions in FLOPs. But on hardware where power is the limiting factor, the usual strategy of just slimming down the model may not work very well. In these cases, this study emphasizes adapting the way a model is used rather than altering the model itself. It’s more about managing the system’s energy and time constraints more efficiently instead of reducing FLOPs.

Models like PP-YOLOE \cite{b13} and EfficientDet \cite{b14} have already shown us that we can gain speed by optimizing internal components. But our approach is different. Instead of redesigning the architecture or requiring retraining, we add a lightweight control layer at inference time. It can sit on top of existing models without needing deep changes. From a deployment standpoint, that’s a significant practical advantage.

The practicality of our hardware-aware approach becomes even more relevant in the broader context of local AI. The same system-level bottlenecks that affect object detection also impact the performance of on-device Large Language Models (LLMs) and other generative AI tasks. These models are often constrained not by raw computational power, but by memory bandwidth and data transfer speeds. Our method of dynamically managing computational load offers a valuable paradigm for these scenarios. By developing lightweight, adaptive runtimes, we can significantly improve the responsiveness and usability of local AI applications, making powerful models more accessible on everyday consumer hardware without compromising user experience.

Its practicality becomes even more relevant when we consider recent trends. Recent models like YOLO-World \cite{b15} and RT-DETR \cite{b16} push toward generalization and flexibility, with architecture designed to scale up across tasks and vocabularies. Yet these models often rely on high-end GPUs and large training budgets. In contrast, our method works with static, pre-trained models and runs effectively on edge devices with tight resource limits. 

\section{\textbf{Conclusion}}
We introduced a Two-Pass Adaptive Inference pipeline for YOLOv10s that speeds up inference on lower power hardware without changing the model architecture. The basic idea is straightforward: we use a lightweight, confidence-based routing system to dynamically tweak the input resolution during PyTorch inference. Testing this on an NVIDIA RTX 4060 Laptop GPU, we observed an 85\% speed gain in our experiments compared to an architectural early-exit baseline, with a mAP loss of only 5.51\% on the COCO 2017 validation set. This work demonstrates that focusing on adaptive runtime strategies, rather than solely on model optimization, is a highly effective path toward enabling high-performance, real-time AI on the devices people use every day.

This suggests a shift in how we might think about deployment. Instead of constantly trying to squeeze out more efficiency by tweaking model architectures alone, it may be more practical to build systems that are aware of the hardware they’re running on. That includes things like dynamic inference strategies and avoiding cloud dependency when privacy matters. As on-device AI, including LLMs, becomes more prevalent, such intelligent, hardware-aware systems will be crucial for delivering a seamless user experience.

However, there are limits. Our tests were done on just one type of GPU and one object detection model (YOLOv10s). Future work should probably explore how this method performs on other hardware including ARM-based SoCs, edge TPUs, and so on with newer versions of YOLO. Additionally, more adaptive thresholds or reinforcement-learned fallback policies might improve the system even further. Our findings point toward a model where intelligent systems adapt not just to data, but also to the devices they run on.


\begin{thebibliography}{00}

\bibitem{b1} 
Wang, A., Chen, H., Liu, L., Chen, K., Lin, Z., Han, J., \& Ding, G. (2024). 
YOLOv10: Real-time end-to-end object detection. 
\textit{arXiv preprint arXiv:2405.14458}. 
https://arxiv.org/abs/2405.14458

\bibitem{b2} 
Rahmath, H. P., Srivastava, V., Chaurasia, K., Pacheco, R. G., \& Couto, R. S. (2024). 
Early-exit deep neural network: A comprehensive survey. 
\textit{ACM Computing Surveys, 57}(3), Article 75. 
https://doi.org/10.1145/3698767

\bibitem{b3} 
Yang, L., Han, Y., Chen, X., Song, S., Dai, J., \& Huang, G. (2020). 
Resolution adaptive networks for efficient inference. 
In \textit{Proceedings of the IEEE/CVF Conference on Computer Vision and Pattern Recognition (CVPR)} (pp. 2366--2375). 
https://doi.org/10.1109/CVPR42600.2020.00244

\bibitem{b4} 
Masum, M. I., Riggs, H., Dey, P., Boymelgreen, A., \& Sarwat, A. I. (2025). 
YOLOv5 vs. YOLOv8 in marine fisheries: Balancing class detection and instance count. 
In \textit{Proceedings of IEEE SoutheastCon} (pp. 607--612). Concord, NC, USA. 
https://doi.org/10.1109/SoutheastCon56624.2025.10971608

\bibitem{b5} 
Kariri, A., \& Elleithy, K. (2025). 
AEA-YOLO: Adaptive enhancement algorithm for challenging environment object detection. 
\textit{AI, 6}(7), Article 132. 
https://doi.org/10.3390/ai6070132

\bibitem{b6} 
Muzammul, M., Li, X., \& Li, X. (2025). 
Enhancing tiny object detection using guided object inference slicing (GOIS). 
\textit{Neurocomputing, 640}, Article 130327. 
https://doi.org/10.1016/j.neucom.2025.130327

\bibitem{b7} 
Edozie, E., Shuaibu, A. N., John, U. K., \& Sadiq, B. O. (2025). 
Comprehensive review of visual object detection. 
\textit{Artificial Intelligence Review, 58}, Article 277. 
https://doi.org/10.1007/s10462-025-11284-w

\bibitem{b8} 
Zhou, H.-Y., Gao, B.-B., \& Wu, J. (2017). 
Adaptive feeding: Achieving fast and accurate detections by adaptively combining object detectors. 
In \textit{Proceedings of the IEEE International Conference on Computer Vision (ICCV)} (pp. 3525--3533). 
https://doi.org/10.1109/ICCV.2017.379

\bibitem{b9} 
Han, Y., Huang, G., Song, S., Yang, L., Wang, H., \& Wang, Y. (2022). 
Dynamic neural networks: A survey. 
\textit{IEEE Transactions on Pattern Analysis and Machine Intelligence, 44}(11), 7436--7456. 
https://doi.org/10.1109/TPAMI.2021.3117837

\bibitem{b10} 
Choquette, J., Gandhi, W., Giroux, O., Stam, N., \& Krashinsky, R. (2021). 
NVIDIA A100 Tensor Core GPU: Performance and innovation. 
\textit{IEEE Micro, 41}(2), 29--35. 
https://doi.org/10.1109/MM.2021.3061394

\bibitem{b11} 
Jiang, Y., Wang, Y., Zhao, M., Zhang, Y., \& Qi, H. (2025). 
Nighttime traffic object detection via adaptively integrating event and frame domains. 
\textit{Fundamental Research, 5}(4), 1633--1644. 
https://doi.org/10.1016/j.fmre.2023.08.004

\bibitem{b12} 
Terven, J., C\'ordova-Esparza, D.-M., \& Romero-Gonz\'alez, J.-A. (2023). 
A comprehensive review of YOLO architectures in computer vision: From YOLOv1 to YOLOv8 and YOLO-NAS. 
\textit{Machine Learning and Knowledge Extraction, 5}(4), Article 83. 
https://doi.org/10.3390/make5040083

\bibitem{b13} 
Xu, S., Wang, X., Lv, W., \& Chang, Q. (2022). 
PP-YOLOE: An evolved version of YOLO. 
\textit{arXiv preprint arXiv:2203.16250}. 
https://arxiv.org/abs/2203.16250

\bibitem{b14} 
Tan, M., Pang, R., \& Le, Q. V. (2020). 
EfficientDet: Scalable and efficient object detection. 
In \textit{Proceedings of the IEEE/CVF Conference on Computer Vision and Pattern Recognition (CVPR)} (pp. 10781--10790). 
https://doi.org/10.1109/CVPR42600.2020.01079

\bibitem{b15} 
Cheng, T., Song, L., Ge, Y., Liu, W., Wang, X., \& Shan, Y. (2024). 
YOLO-World: Real-time open-vocabulary object detection. 
In \textit{Proceedings of the IEEE/CVF Conference on Computer Vision and Pattern Recognition (CVPR)} (pp. 16901--16911). 
https://doi.org/10.1109/CVPR52733.2024.01599

\bibitem{b16} 
Zhao, Y., Lv, W., Xu, S., Wei, J., Wang, G., Dang, Q., Liu, Y., \& Chen, J. (2024). 
DETRs beat YOLOs on real-time object detection. 
In \textit{Proceedings of the IEEE/CVF Conference on Computer Vision and Pattern Recognition (CVPR)} (pp. 16965--16974). 
https://doi.org/10.1109/CVPR52733.2024.01605

\end{thebibliography}
\end{document}